\documentclass[final]{elsarticle}
\usepackage{hyperref}
\usepackage{verbatim}
\usepackage{booktabs}

\begin{document}

\begin{frontmatter}

\title{Inference of visual field test performance from OCT volumes using deep learning}

\author[mymainaddress]{Stefan Maetschke}
\author[mymainaddress]{Bhavna Antony}
\author[mysecondaryaddress]{Hiroshi Ishikawa}
\author[mysecondaryaddress]{Gadi Wollstein}
\author[mysecondaryaddress]{Joel S. Schuman}
\author[mymainaddress]{Rahil Garnavi}

\address[mymainaddress]{IBM Research Australia, Melbourne, VIC, Australia}
\address[mysecondaryaddress]{NYU Langone Eye Center, New York University School of Medicine, New York, NY, USA.}

\begin{abstract}
Visual field tests (VFT) are pivotal for glaucoma diagnosis and conducted regularly to monitor disease progression.
Here we address the question to what degree aggregate VFT measurements such as Visual Field Index (VFI) and Mean Deviation (MD) can be inferred from Optical Coherence Tomography (OCT) scans of the Optic Nerve Head (ONH) or the macula. Accurate inference of VFT measurements from OCT could reduce examination time and cost.

We propose a novel 3D Convolutional Neural Network (CNN) for this task and compare its accuracy with classical machine learning  (ML) algorithms trained on common, segmentation-based OCT, features employed for glaucoma diagnostics.

Peak accuracies were achieved on ONH scans when inferring VFI with a Pearson Correlation (PC) of 0.88$\pm$0.035 for the CNN and a significantly lower (p $<$ 0.01) PC of 0.74$\pm$0.090 for the best performing, classical ML algorithm -- a Random Forest regressor. Estimation of MD was equally accurate with a PC of 0.88$\pm$0.023 on ONH scans for the CNN.
\end{abstract}

\begin{keyword}
Glaucoma, Optical Coherence Tomography, Visual Field Test, Deep Learning
\end{keyword}

\end{frontmatter}


\section{Introduction}

Glaucoma is a progressive and irreversible eye disease, where vision is lost due to damage to the optic nerve. Vision loss is usually gradual, starting peripherally and since there are no symptoms in the early stages, detection of the condition tends to occur late. The disease can be detected via a variety of tests such as planimetry, pachymetry, tonometry, fundus photography and optical coherence tomography (OCT)~\cite{NICE2017} but final diagnosis typically includes a visual field test (VFT)~\cite{Broadway2012}. 

VFTs, however, are costly, time-consuming and media opacities such as cataracts, visual acuity, glaucoma medications, severity of glaucoma, learning effect, distraction and other factors can affect the reliability  of VFTs~\cite{Ho2013,Peracha2013,Longmuir2009}. In addition, agreement between clinicians about visual field progression, even for reliable VFTs, was found to be low ($\kappa=0.32$)~\cite{Viswanathan2003} due to a lack of common consensus on the glaucoma progression assessment criteria. 

On the other hand, OCT~\cite{Huang1991} is a non-invasive imaging modality that enables the objective quantification of retinal structures, which have been found to be important biomarkers of glaucoma. Specifically, the retinal nerve fiber layer (RNFL) and combined ganglion cell with inner plexiform layer (GCIPL) begin to thin early as the disease progresses~\cite{Medeiros2009,Lucy2016}. 

This raises the obvious question to what degree structure and function are correlated and whether functional measurements such as  Visual Field Index (VFI) or Mean Deviation (MD) can be inferred from structure. Harwerth et al.~\cite{Harwerth2006} and Leite at al.~\cite{Leite2012} found visual field defects to be proportional to the neural loss caused by glaucoma. Similarily, Hood et al.~\cite{Hood2011} and Ali et al.~\cite{Raza2011} demonstrated that the visual field thresholds can be directly compared to the OCT probability or thickness maps when correcting for the displacement of the retinal ganglion cells.
An earlier study by Sugimoto et al.~\cite{Sugimoto2013} used a Random Forest classifier to predict glaucomatous VF damage based on age, gender, right or left eye, axial length and 237 different OCT measurements. Later Bogunović et al.~\cite{Bogunovic2015,Abramoff2015} and Guo et al~\cite{Guo2017} estimated individual Humphrey 24-2 visual field thresholds from RNFL and GCIPL thickness maps derived from nine-field SD-OCT using Support Vector Machines.

Here we assess whether aggregate VFT metrics such as VFI and MD can be estimated directly from single, raw OCT scans of the ONH or the macula. Thus eliminating the need for 9-field OCT or layer segmentation, which is prone to errors especially with moderate to severe glaucoma conditions and also co-existing other ocular pathologies. We compare the accuracy of the proposed Convolutional Neural Network (CNN)~\cite{Krizhevsky2012,Lecun2015} with classical approaches that apply machine learning (e.g. k-Nearest Neighbor, Support Vector Machines, Random Forests and others~\cite{Russell2010}) to segmentation-based OCT features such as peripapillary RNFL thickness at 12 clock-hours and four quadrants, mean retinal nerve fiber layer (RNFL) thickness, rim and disc area, horizontal/vertical cup-to-disc ratio and cup volume.

\section{Material and methods}

This study was an observational study that was conducted in accordance with the tenets of the Declaration of Helsinki and the Healthy Insurance Portability and Accountability Act. The Institutional Review Board of New York University and the University of Pittsburgh approved the study, and all subjects gave written consent before participation.

\subsection{Performance metrics}

Two common, global Visual Field Test (VFT) indices employed for glaucoma diagnostics are Mean Deviation (MD) and Visual Field Index (VFI). The MD measures the reduction in sensitivity, averaged across the visual field and relative to a group of healthy, age-matched patients. The VFI is calculated from the pattern deviation probability plot with a MD~$\geq$~-20 dB or from the total deviation probability plot for a MD~$<$~-20 dB. The VFI ranges from 100\% (healthy) to 0\% (perimetrically blind) \cite{Bengtsson2008}. Note that VFI and MD are highly correlated ($r=-0.96$) but the VFI has reduced sensitivity to early damage~\cite{Mansuri2014}.

The accuracy of the evaluated deep and machine learning methods to estimate VFI and MD was quantified by the Root Mean Square Error ($RMSE$) and the Pearson Correlation ($PC$). The $RMSE$ is defined as follows
$$
RMSE = \frac{1}{n} \sqrt{\sum_{i}^{n} (x_i-y_i)^2} \ , 
$$
where $x_i$ is the estimated and $y_i$ is the true, measured VFI or MD of the $i$-th visual field test. Similarly, the correlation coefficient $PC$ was computed as
$$
PC = \frac{\sum_{i}^{n} (x_i-\bar{x}) (y_i-\bar{y})}
{\sqrt{\sum_{i}^{n} (x_i-\bar{x})^2 \sum_{i}^{n} (y_i-\bar{y})^2}} \ , 
$$
where $\bar{x}$ and $\bar{y}$ are the samples means.

\subsection{Data}

ONH and macula scans from 579 patients were acquired on a Cirrus HD-OCT Scanner (Zeiss, Dublin, CA, USA). 504 of the 4155 scans were diagnosed as healthy and 3651 with primary open angle glaucoma (POAG). The data set was divided according to a ratio of 80\% training samples, 10\% validation samples and 10\% test samples. It was ensured that scans belonging to the same patient were not scattered across folds. 

Scans were labeled as glaucoma if glaucomatous visual field defects were present and at least 2 consecutive test results were abnormal. The scans had physical dimensions of 6x6x2~mm with a corresponding size of 200x200x1024~voxels per volume but were down-sampled to 64x64x128 for network training and normalized with respect to laterality (left eyes were flipped to right eyes).

Corresponding visual field test measurements (MD and VFI) were aquired on a Zeiss Meditec, Model 750, with test type SITA Standard 24-2. Following \cite{Yaqub2012}, tests with pupil diameter less than 2.5mm, fixation loss greater than 20\% or false positive or negative rates of more than 20\% were discarded.

Demographical background such as gender and race distribution, and mean values with standard deviations for patient's age, Intraocular Pressure (IOP), MD and VFI measurements are provided in Table~\ref{tab:demo}. Statistically significant differences ($p < 0.0001$) between the distribution of healthy and POAG patients were found for age, IOP, MD and VFI.

\begin{table}[!ht]
	\centering
	\caption{{\bf Demographic data: Gender and race distribution, and mean values with standard deviations and ranges for age, IOP, MD and VIF.}}
	\begin{tabular}{lcc}
		& Healthy & POAG \\
		\hline \noalign{\vskip 3pt}
		\#Female 	& 76				& 246 \\
		\#Male 		& 39				& 217 \\
		\#Uknown    & 0				    & 1 \\
		\hline \noalign{\vskip 3pt} 
		\#White 	& 87				& 308 \\  
		\#Black 	& 26				& 142 \\ 
		\#Asian 	& 0					& 12 \\ 
	    \#Hispanic 	& 0					& 1 \\ 
	    \#Unknown 	& 2					& 1 \\ 
		\hline \noalign{\vskip 3pt}
		Age 		& 60.2$\pm$15.3 [22.8..87.0]	& 66.2$\pm$12.5 [25.2..93.3]\\
		\hline \noalign{\vskip 3pt} 
		IOP 		& 13.6$\pm$2.7	[8..23]			& 15.3$\pm$4.6 [1..60] \\ 
		\hline \noalign{\vskip 3pt}
		MD 			& -0.74$\pm$1.5 [-10.7..2.8]	& -7.2$\pm$8.1 [-32.5..2.8]\\
		\hline \noalign{\vskip 3pt}
		VFI 		& 98.7$\pm$2.0 [77..100]		& 80.7$\pm$25.1 [0..100]\\		
		\hline 
	\end{tabular}
	\label{tab:demo}
\end{table}

\subsection{Classical machine learning}

Traditionally, ONH related features such as RNFL thickness, cup-to-disc ratio and others are measured on OCT images and build the basis for glaucoma assessment. We use these numerical features as input to a wide range of classical machine learning algorithms but aim to estimate VFT metrics (VFI, MD) instead of detecting glaucoma. Note that this is a regression problem in contrast to a classification problem.

Specifically, we take 22 features extracted by the Cirrus OCT scanner such as peripapillary RNFL thickness at 12 clock-hours, peripapillary RNFL thickness in the four quadrants, average RNFL thickness, rim area, disc area, average cup-to-disc ratio, vertical cup-to-disc ratio and cup volume~\cite{Kim2009}. 

All features were normalized by subtracting the features mean and scaling to unit variance. Normalization parameters were estimated on the training data only and then applied to training, validation and test data. No further pre-processing steps were performed. All features were real valued and contained no missing values. 

We then trained the following machine learning algorithms as implemented in the Scikit-learn library~\cite{Pedregosa2011} on the extracted 22 features: Linear Regression and k-nearest Neighbor Regression~\cite{Russell2010},  Support Vector Regression (linear, polynomial, RBF)~\cite{Scholkopf2001}, Multi-layer Perceptron~\cite{Rumelhart1986}, Random Forest Regression and Decision Tree Regression~\cite{Breiman2001}, Gradient Boosting Regression~\cite{Natekin2013} and Extra Trees Regression~\cite{Geurts2006}. 

Hyper-parameters of each algorithm were optimized as follows: we selected important hyper-parameters with reasonable ranges, and then uniformly sampled 100 times for each training fold. The parameters resulting in the highest PC on the validation set were used to compute the PC and RMSE on the test set. This process was repeated 5 times (5-fold cross-validation) and we report mean PC and RMSE with standard deviations (STD) on the test set.

\subsection{Deep learning}

In contrast to the classical machine learning approach the proposed deep learning system does not extract a predefined set of glaucoma related features such as RNFL thickness and other from the OCT volume but operates on the raw data. Apart from down-sampling (linear interpolation) from 200x200x1024 to volumes with dimensions 64x64x128 voxels due to constraints of the GPU memory (12GB), and flipping of eyes, no further pre-processing or feature extraction was performed.

\begin{figure}[h]
	\centering
	\includegraphics[width=1.0\linewidth]{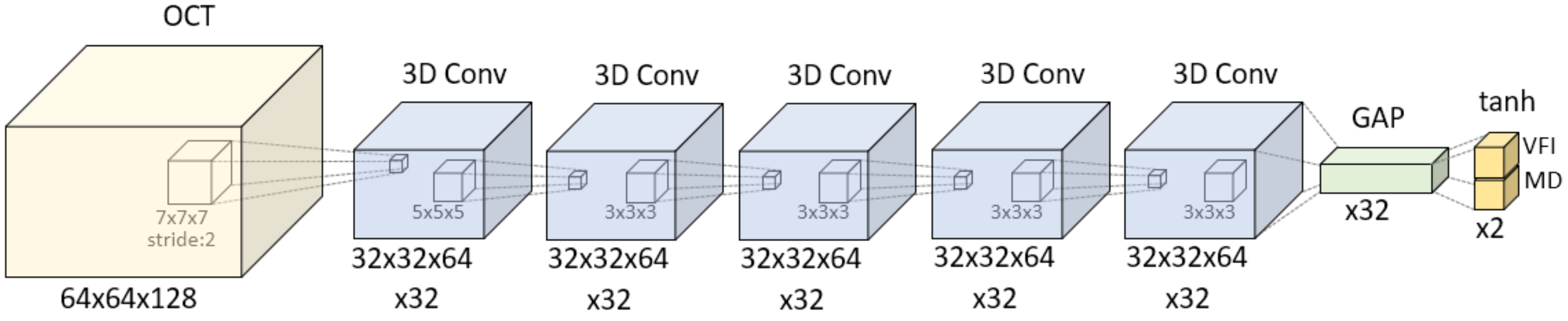}
	\caption{Network architecture.}
	\label{fig:architecture}
\end{figure}

The downsampled volumes were inputted into a Convolutional Neural Network (CNN)~\cite{Lecun2015}, depicted in Figure~\ref{fig:architecture}, where the network architecture is derived from a CNN for glaucoma classification~\cite{Maetschke2018}. It is composed of five 3D-convolutional layers with ReLU activation, batch-normalization, filter banks of sizes 32-32-32-32-32, filters of sizes 7-5-3-3-3 and strides 2-1-1-1-1. We also employed 3D spatial dropout ($rate=0.2$)~\cite{Tompson2015,Chollet2017} to counteract over-fitting. In contrast to \cite{Maetschke2018}, after the Global Average Pooling (GAP)~\cite{Zhou2015} a regression layer with Tangens hyperbolicus (tanh) activation function is added to infer VFI and MD. Note that 3D convolutions are essential to enable the computation of 3D Class Activation Maps (CAM)~\cite{Zhou2015}. 

The CNN was implemented in Keras~\cite{Chollet2017} with Tensorflow~\cite{Abadi2016} as backend. Data splitting and pre-processing was performed with nuts-flow/ml~\cite{Maetschke2017}. Training was performed on a single K80 GPU using NAdam with a learning rate of $1e-4$ over 300 epochs and 5-fold cross-validation. The network with the highest validation PC during training was saved (early stopping) and results on the independent test are reported.
 
CAMs were constructed following Zhou et al.~\cite{Zhou2015}, resized and overlayed on the input OCT scan. Note that CAMs are computed for the smaller input OCTs 64x64x128 and mapped back to scans with the original dimensions of 200x200x1024 (See Figure~\ref{fig:cams}).

\section{Results}

In this section we first report the results for the classical machine learning methods trained on segmentation-based OCT features, and then the performance of the deep learning system trained on raw OCT scans of the ONH or the macula.

Table~\ref{tab:ml-results} shows the Person Correlation (PC) between the estimated and the true VFI of the various machine learning methods on the test data. As stated above, training and testing was performed via 5-fold cross-validation and the reported mean PCs and standard deviations are computed over the 5 test folds.

\begin{table}[!ht]
	\centering
	\begin{tabular}{lccc}
		\toprule
		Method  & PC:VFI \\
		\midrule
		RFR                  & 0.74$\pm$0.090\\
		GBR                  & 0.72$\pm$0.087\\		
		SVR (RBF)            & 0.71$\pm$0.105\\		
		LR                   & 0.69$\pm$0.075\\	
		KNR                  & 0.69$\pm$0.075\\		
		SVR (linear)         & 0.66$\pm$0.060\\			
		MLP                  & 0.64$\pm$0.113\\		
		DTR                  & 0.62$\pm$0.084\\	
		ETR                  & 0.59$\pm$0.089\\	
		SVR (poly)           & 0.42$\pm$0.112\\				
		\bottomrule
	\end{tabular}
	\caption{Results for classical machine learning methods such as Random Forest Regression (RFR), Gradient Boosting Regression (GBR), Support Vector Regression (SVR) with linear, polynomial or RBF kernel, Linear Regression (LR), k-nearest Neighbor Regression (KNR) ,  Multilayer Perceptron (MLP), Decision Tree Regression (DTR) and Extra Tree Regression (ETR). Performance is measured by Person Correlation (PC) for Visual Field Index (VFI) with standard deviations over 5-fold cross-validation. }	
	\label{tab:ml-results}
\end{table}

The highest correlation of $0.74$ was achieved by the Random Forest Regressor (RFR), closely followed by Support Vector Regression with RBF kernel and Gradient Boosting Regression (GBR). These non-linear regression methods outperformed the standard Linear Regression (LR). Support Vector Machine Regression with linear or polynomial kernel was found inferior and comparatively time consuming to train. Similarly, decision tree regressors (DTR, ETR), while fast to train, did not achieve high accuracies.

The results of the deep learning system trained on raw OCT scans of the ONH or macula are shown in Table~\ref{tab:dl-results}. In addition to the PC, here we also report the RMSE between the inferred and the true VFI or MD. As before, all results are 5-fold cross-validated.

\begin{table}[!htbp]
	\centering
	\begin{tabular}{ccccc}
		\toprule
		{} &  \multicolumn{2}{c}{PC} & \multicolumn{2}{c}{RMSE}\\
		\midrule
		Region  & VFI & MD & VFI & MD \\
	    \midrule		
		ONH 		& 0.88$\pm$0.029	& 0.88$\pm$0.024	& 12.0$\pm$1.61	& 4.1$\pm$0.44 \\
		Macula 		& 0.86$\pm$0.026	& 0.85$\pm$0.016	& 13.4$\pm$1.75	& 4.4$\pm$0.38 \\
		\bottomrule
	\end{tabular}
	\caption{Results of the deep learning network for ONH and Macula scans. Accuracy is measured by Root Mean Square Error (RMSE) and Pearson Correlation (PC) of Mean Deviation (MD) and Visual Field Index (VFI) with standard deviations over 5-fold cross-validation. }
    \label{tab:dl-results}
\end{table}

In comparison with the classical ML methods trained on traditional OCT features the CNN achieved a substantially higher correlation of $0.88$. The PC on Macula scans was slightly lower ($0.86$) but still significantly higher than the PC of $0.74$ for the best classical ML method (RFR). Note that the reported RMSE for VFI and MD are not directly comparable, due to the different ranges of the metricies (VFI: $0..100$, MD: $-32.5..2.8$).  The PC shows no significant difference between VFI and MD for the same region (ONH or macula).

\begin{figure}[h!]
	\centering
	\includegraphics[width=0.9\linewidth]{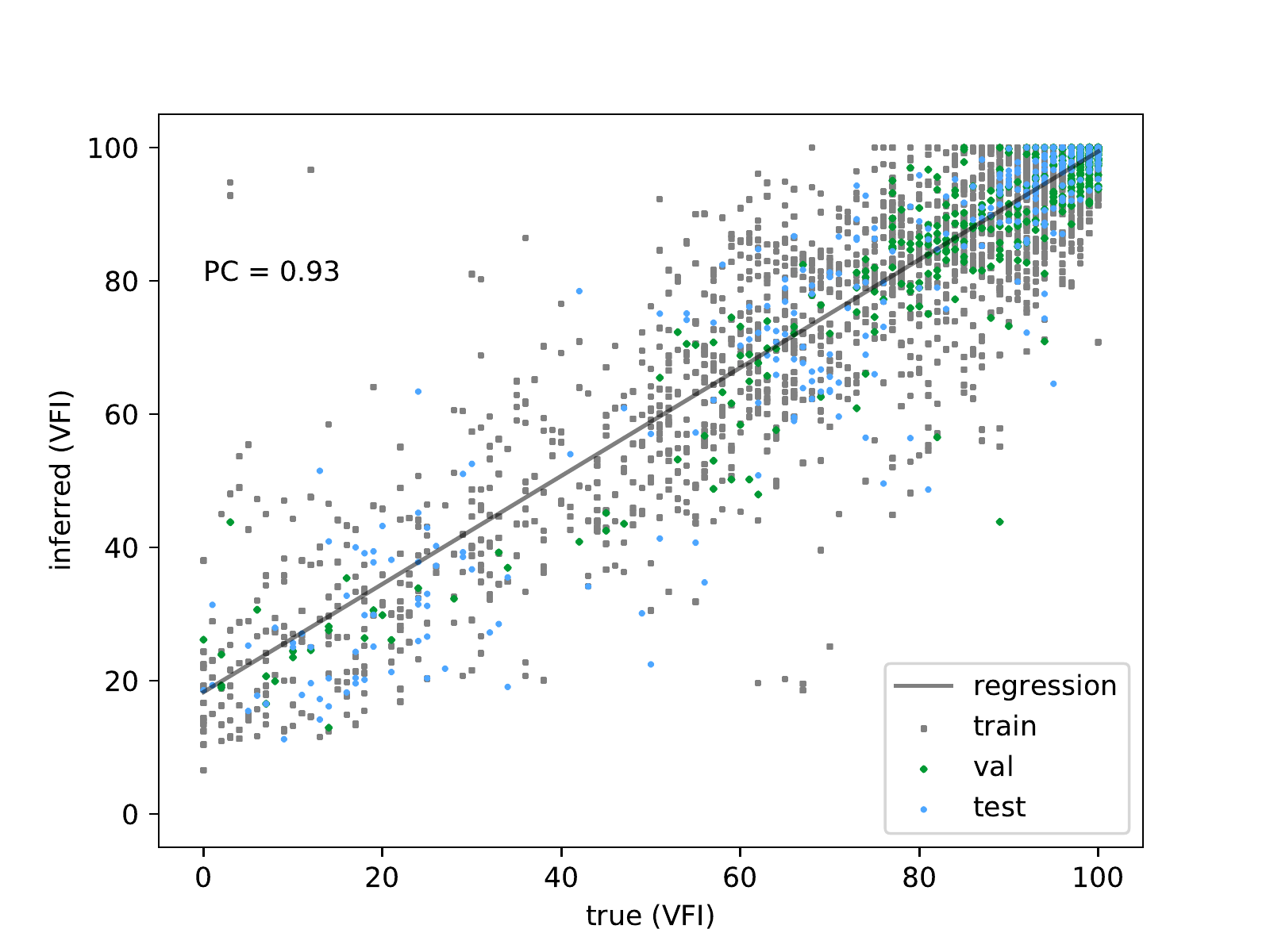}
	\caption{Scatter plot of estimated versus true VFI for a training, validation and test fold for the deep learning network trained on ONH. Black line shows linear regression with PC of $0.93$. (Plot best seen in color) }
	\label{fig:pred_scatter}
\end{figure}

Figure~\ref{fig:pred_scatter} provides an overview of the CNN's VFI estimates by plotting the inferred versus the true VFI for a typical training, validation and test fold. Note that the regression line is computed on the entire data set (4155 samples) and the corresponding PC of $0.93$ is therefore higher than the reported mean PC of $0.88$ on the test folds. The data set shows an expected bias towards mild cases of glaucoma (VFI $>$ 80) and there are a few cases where the system heavily under- or over-estimates the VFI.

\subsection{Class Activation Maps}

\begin{figure}[!h]
	\centering
	\includegraphics[width=0.5\linewidth]{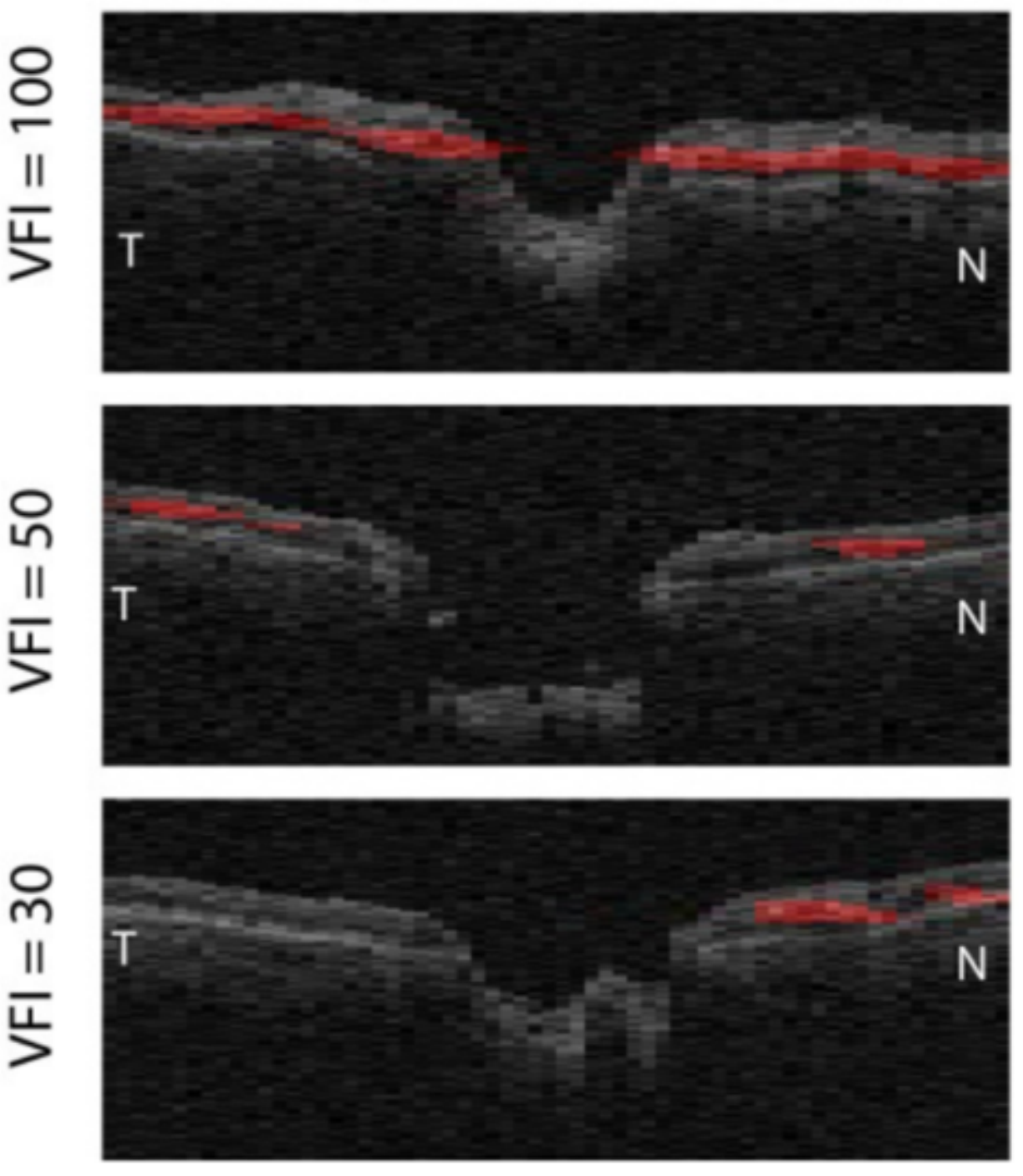}
	\caption{Class activation maps for three different VFI values. Regions and layers important for VFI inference are marked red.
	N=Nasal, T=Temporal. (Maps best seen in color)}
	\label{fig:cams}
\end{figure}

In attempt to understand which regions of an OCT scan are informative for the estimation of VFI, we computed Class Activation Maps (CAMs)~\cite{Zhou2015}. Figure~\ref{fig:cams} shows representative CAMs for three patients with different degrees of vision loss measured by the VFI. Note that aspect ratios of scans do not reflect physical dimensions of OCT volumes.

In the case of normal vision (VFI = 100) the network focuses on the Ganglion Cell and Inner Plexiform layers (GCIPL) in general (marked red). For increasingly higher degrees of vision loss (VFI = 50, VFI=30), smaller more specific regions of the GCIPL are highlighted; potentially indicating locations of layer thinning.

\section{Discussion}

The deep learning network trained on raw OCT scans of the ONH or macula achieved significantly higher accuracies in estimating VFI or MD than any of the classical machine learning methods trained on segmentation-based OCT features. While the computation time for estimates of both approaches is comparable, the CNN is substantially more time-consuming to train; days in contrast to minutes. Furthermore, while CAMs provide some insight into the features used by the CNN, the explanatory power of the established features  RNFL thickness, cup-to-disc ratio and others remains better.

We attempted to further improve the accuracy of the deep learning system by augmenting the OCT volumes during training. Specifically, we explored random occlusions, translations, small rotations ($\pm$10 degrees) along the enface axis, and mixup~\cite{Zhang2017} but did not observe a significant impact on accuracy and therefore did not employ augmentation for the final results reported here. 

Other variation such as higher-resolution OCT volumes or combining ONH and macula scans, which are likely to improve the accuracy of the CNN were not evaluated. It is noteworthy that we also did not use Intraoccular Pressure (IOP) or age as input features, where the latter is a required parameter for VFT, as threshold values are compared to age-adjusted normal values. 

VFI and MD measurements provide the ground truth data the CNN was trained on. These aggregate measures are derived from VFT, which are known to be unreliable and perfect accuracy can therefore not be expected. Abramoff et al.~\cite{Abramoff2015} computed the retest variability of VTFs, which defines an upper bound on the accuracy of the CNN, and found a correlation of $0.88$. This might indicate that the performance of the CNN, with a PC of $0.88$ is close too or at the upper bound. Also in very good agreement with \cite{Abramoff2015} is the accuracy of the SVM for which Abramoff et al. report a mean correlation of $0.68$, while our evaluation finds a correlation of $0.66$ (see SVM(linear)in Table~\ref{tab:ml-results}).

Guo et al.\cite{Guo2017} and  Bogunović et al.\cite{Bogunovic2015} estimate individual Humphrey 24-2 visual field thresholds but require segmentation of RNFL and GCL, and the registration of the 9-field OCT volumes. The potential errors in registration and segmentation, especially for severe cases of glaucoma or lower quality scans, might have a negative effect on their estimates. Furthermore, Raza et al.~\cite{Raza2011} found the RNFL of the temporal region of the disc to be thin and highly variable, even in healthy individuals, indicating that RNFL thickness might not be a good measure of structural damage. In contrast, the method proposed here avoids explicit segmentation and registration but is currently limited to estimating aggregate VFT indices such as VFI and MD.

\section{Conclusions}

We demonstrated that aggregate VFT measurements such as VFI and MD for healthy and POAG patients can be estimated directly and with good accuracy from raw OCT scans using a Convolutional Neural Network. We found the accuracy when using ONH scans as inputs slightly higher than for macula scans. The deep learning approach achieved considerably higher accuracies than all other machine learning methods trained on the features usually employed for glaucoma diagnosis. An method that accurately infers VFT measurements from OCT scans has the potential of substantially reducing examination time and cost.

\bibliography{mybibfile}

\end{document}